# Error-tolerant Finite State Recognition with Applications to Morphological Analysis and Spelling Correction

Kemal Oflazer*
Bilkent University

*This paper presents a notion of error-tolerant recognition with finite state recognizers along with results from some applications. Error-tolerant recognition enables the recognition of strings that deviate mildly from any string in the regular set recognized by the underlying finite state recognizer. Such recognition has applications in error-tolerant morphological processing, spelling correction, and approximate string matching in information retrieval. After a description of the concepts and algorithms involved, we give examples from two applications: In the context of morphological analysis, error-tolerant recognition allows misspelled input word forms to be corrected, and morphologically analyzed concurrently. We present an application of this to error-tolerant analysis of agglutinative morphology of Turkish words. The algorithm can be applied to morphological analysis of any language whose morphology is fully captured by a single (and possibly very large) finite state transducer, regardless of the word formation processes and morphographemic phenomena involved. In the context of spelling correction, error-tolerant recognition can be used to enumerate correct candidate forms from a given misspelled string within a certain edit distance. Again, it can be applied to any language with a word list comprising all inflected forms, or whose morphology is fully described by a finite state transducer. We present experimental results for spelling correction for a number of languages. These results indicate that such recognition works very efficiently for candidate generation in spelling correction for many European languages such as English, Dutch, French, German, Italian (and others) with very large word lists of root and inflected forms (some containing well over 200,000 forms), generating all candidate solutions within 10 to 45 milliseconds (with edit distance 1) on a SparcStation 10/41. For spelling correction in Turkish, error-tolerant recognition operating with a (circular) recognizer of Turkish words (with about 29,000 states and 119,000 transitions) can generate all candidate words in less than 20 milliseconds, with edit distance 1.*

## 1. Introduction

Error-tolerant finite state recognition enables the recognition of strings that deviate *mildly* from any string in the regular set recognized by the underlying finite state recognizer. For example, suppose we have a recognizer for the regular set over $\{a, b\}$ described by the regular expression $(aba + bab)^*$, and we would like to recognize inputs which may be corrupted (but not too much) due to a number of reasons: e.g., *abaaaba* may be matched to *abaaba* correcting for a spurious *a*, or *babbb* may be matched to *babbab* correcting for a deletion, or *ababba* may be matched to either *abaaba* correcting a *b* to an *a* or to *ababab*

* Department of Computer Engineering and Information Science, Bilkent University, Ankara, TR-06533, Turkey

correcting the reversal of the last two symbols. Error-tolerant recognition can be used in many applications that are based on finite state recognition, such as morphological analysis, spelling correction, or even tagging with finite state models (Voutilainen and Tapanainen, 1993; Roche and Schabes, 1995). The approach presented in this paper uses the same finite state recognizer that is built to recognize the regular set, but relies on a very efficiently controlled recognition algorithm based on depth-first search of the state graph of the recognizer. In morphological analysis, misspelled input word forms can be corrected, and morphologically analyzed concurrently. In the context of spelling correction, error-tolerant recognition can universally be applied to the generation of candidate correct forms, for any language with a word list comprising all inflected forms, or whose morphology is fully described by automata such as two–level finite state transducers (Karttunen and Beesley, 1992; Karttunen, Kaplan, and Zaenen, 1992). The algorithm for error-tolerant recognition is very fast and applicable to languages which have productive compounding, and/or agglutination as word formation processes.

There have been a number of approaches to error-tolerant searching. Wu and Manber (Wu and Manber, 1991) describe an algorithm for fast searching, allowing for errors. This algorithm (called *agrep*) relies on a very efficient pattern matching scheme whose steps can be implemented with a arithmetic and logical operations. It is however efficient when the size of the pattern is limited to 32 to 64 symbols, though it allows for an arbitrary number of insertions, deletions and substitutions. It is very suitable when the pattern is very small and the sequence to be searched is very large. Myers and Miller (Myers and Miller, 1989) describe algorithms for approximate matching to regular expressions with arbitrary costs, but these are again suited for applications where the pattern or the regular expression is small and the sequence is large. Schneider *et al.* (Schneider, Lim, and Shoaff, 1992) present a method for imperfect string recognition using fuzzy logic. Their method is for context-free grammars, hence can be applied to finite state recognition also, but relies on introducing new productions to allow for errors which may increase the size of the grammar substantially.

The paper first presents the notion of error tolerant recognition with finite state recognizers and presents an algorithm for error tolerant recognition with an arbitrary finite state recognizer. It then presents an application of the approach to error-tolerant morphological analysis with transducers, along with an example from Turkish morphology. The paper then presents an application of error-tolerant recognition to candidate generation in spelling correction, along with extensive results from many languages.

## 2. Error-tolerant Finite State Recognition

We can informally define error-tolerant recognition with a finite state recognizer as the recognition of all strings in the regular set (accepted by the recognizer), and additional strings which can be obtained from any string in the set *by a small number of unit editing operations*.

The notion of error-tolerant recognition requires an error metric for measuring how much two strings deviate from each other. The *edit distance* between two strings measures the minimum number of unit editing operations of *insertion, deletion, replacement of a symbol, and transposition of adjacent symbols* (Damerau, 1964), that are necessary to convert one string into another. Let $Z = z_1, z_2, \ldots, z_p$, denote a generic string of $p$ symbols from an alphabet $A$. $Z[j]$ denotes the initial substring of any string $Z$ up to and including the $j^{th}$ symbol. We will use $X$ (of length $m$) to denote the misspelled string, and $Y$ (of length $n$) to denote the string that is a (possibly partial) candidate string. Given two strings $X$ and $Y$, the edit distance $ed(X[m], Y[n])$ computed according to the recurrence below (Du and Chang, 1992), gives the minimum number of unit editing operations to convert one string to the other.

$$
\begin{aligned}
ed(X[i+1], Y[j+1]) &= ed(X[i], Y[j]) && \text{if } x_{i+1} = y_{j+1} \\
&&& \text{(last characters are same)} \\
&= 1 + min\{ed(X[i-1], Y[j-1]), && \text{if both } x_i = y_{j+1} \\
&\qquad ed(X[i+1], Y[j]), && \text{and } x_{i+1} = y_j \\
&\qquad ed(X[i], Y[j+1])\} && \\
&&& \text{(last two characters are} \\
&&& \text{transposed)} \\
&= 1 + min\{ed(X[i], Y[j]), && \text{otherwise} \\
&\qquad ed(X[i+1], Y[j]), && \\
&\qquad ed(X[i], Y[j+1])\} && \\
ed(X[0], Y[j]) &= j && 0 \leq j \leq n \\
ed(X[i], Y[0]) &= i && 0 \leq i \leq m \\
ed(X[-1], Y[j]) &= ed(X[i], Y[-1]) = \max(m, n) && \text{(Boundary definitions)}
\end{aligned}
$$

For example $ed(recoginze, recognize) = 1$, since transposing *i* and *n* in the former string would give the latter. Similarly $ed(sailn, failing) = 3$ as in the former string, one could change the initial *f* to *s*, insert an *i* before the *n*, and insert a *g* at the end to obtain the latter.

A (deterministic) finite state recognizer, $R$ is described by a 5-tuple $R = (Q, A, \delta, q_0, F)$ with $Q$ denoting the set of states, $A$ denoting the input alphabet, $\delta : Q \times A \rightarrow Q$, denoting the state transition function, $q_0 \in Q$ denoting the initial state, and $F \subseteq Q$ denoting the final states (Hopcroft and Ullman, 1979). Let $L \subseteq A^*$ be the regular language accepted by $R$. Given an edit distance error threshold $t > 0$, we define a string $X[m] \notin L$ to be recognized by $R$ with an error at most $t$, if the set

$$C = \{Y[n] \mid Y[n] \in L \text{ and } ed(X[m], Y[n]) \leq t\}$$

is not empty.

### 2.1 An algorithm for error-tolerant recognition

Any finite state recognizer can also be viewed as a directed graph with arcs are labeled with symbols in $A$.[1] Standard finite state recognition corresponds to traversing a path (possibly involving cycles) in the graph of the recognizer, starting from the start node, to one of the final nodes, so that the concatenation of the labels on the arcs along this path matches the input string. For error-tolerant recognition one needs to find *all paths from the start node to one of the final nodes, so that when the labels on the links along a path are concatenated, the resulting string is within a given edit distance threshold t, of the (erroneous) input string.* With $t > 0$, the recognition procedure becomes a search on this graph as shown in Figure 1.

Searching the graph of the recognizer has to be very fast if error-tolerant recognition is to be of any practical use. This means that paths that can lead to no solutions have to be pruned so that the search can be limited to a very small percentage of the search space. Thus we need to make sure that any candidate string that is generated as the search is being performed, does not deviate from certain initial substrings of the erroneous string by more than the allowed threshold. To detect such cases, we use the notion of a *cut-off*

1 We may use state and node, and transition and arc, interchangeably.

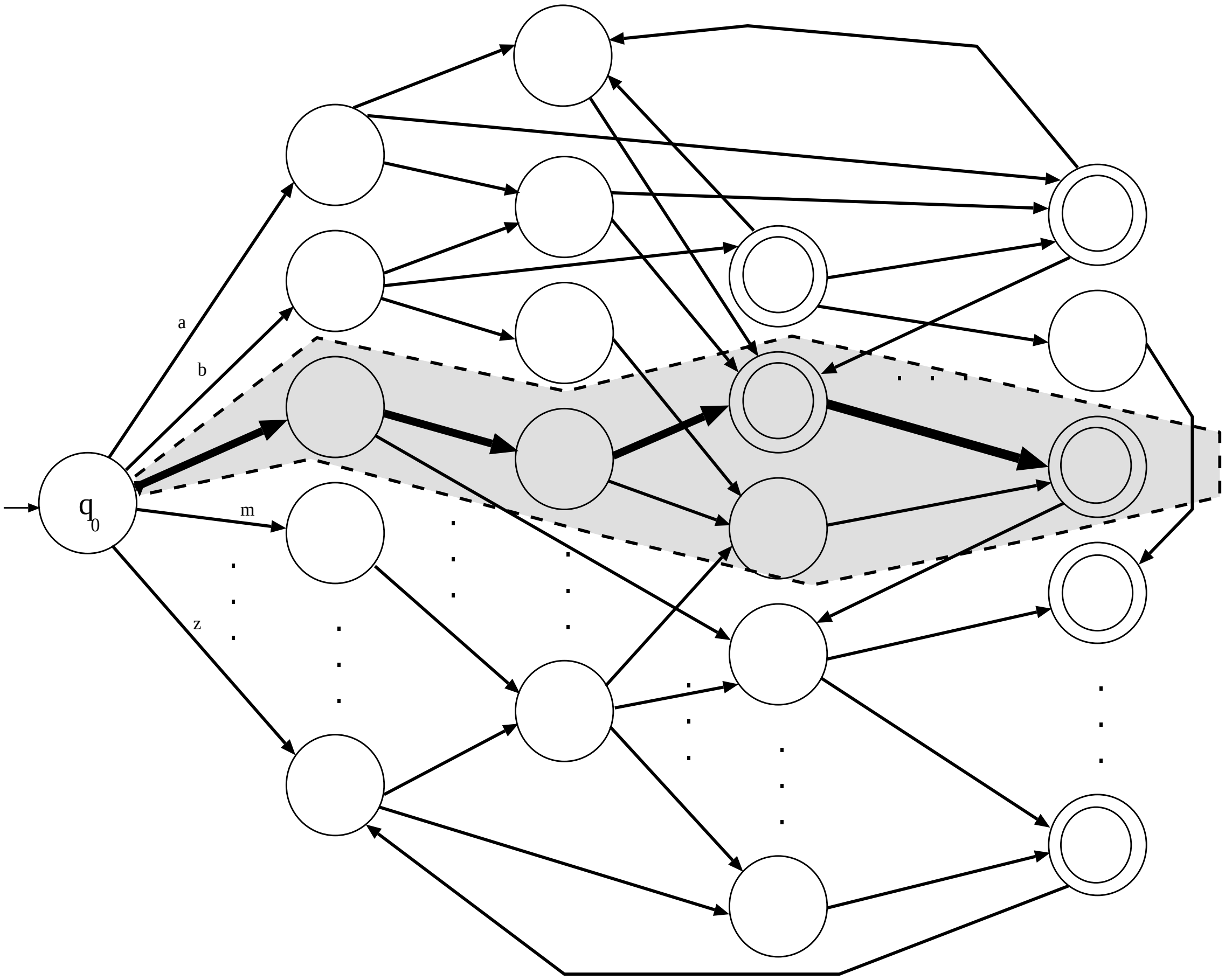

**Figure 1**
Searching the recognizer graph.

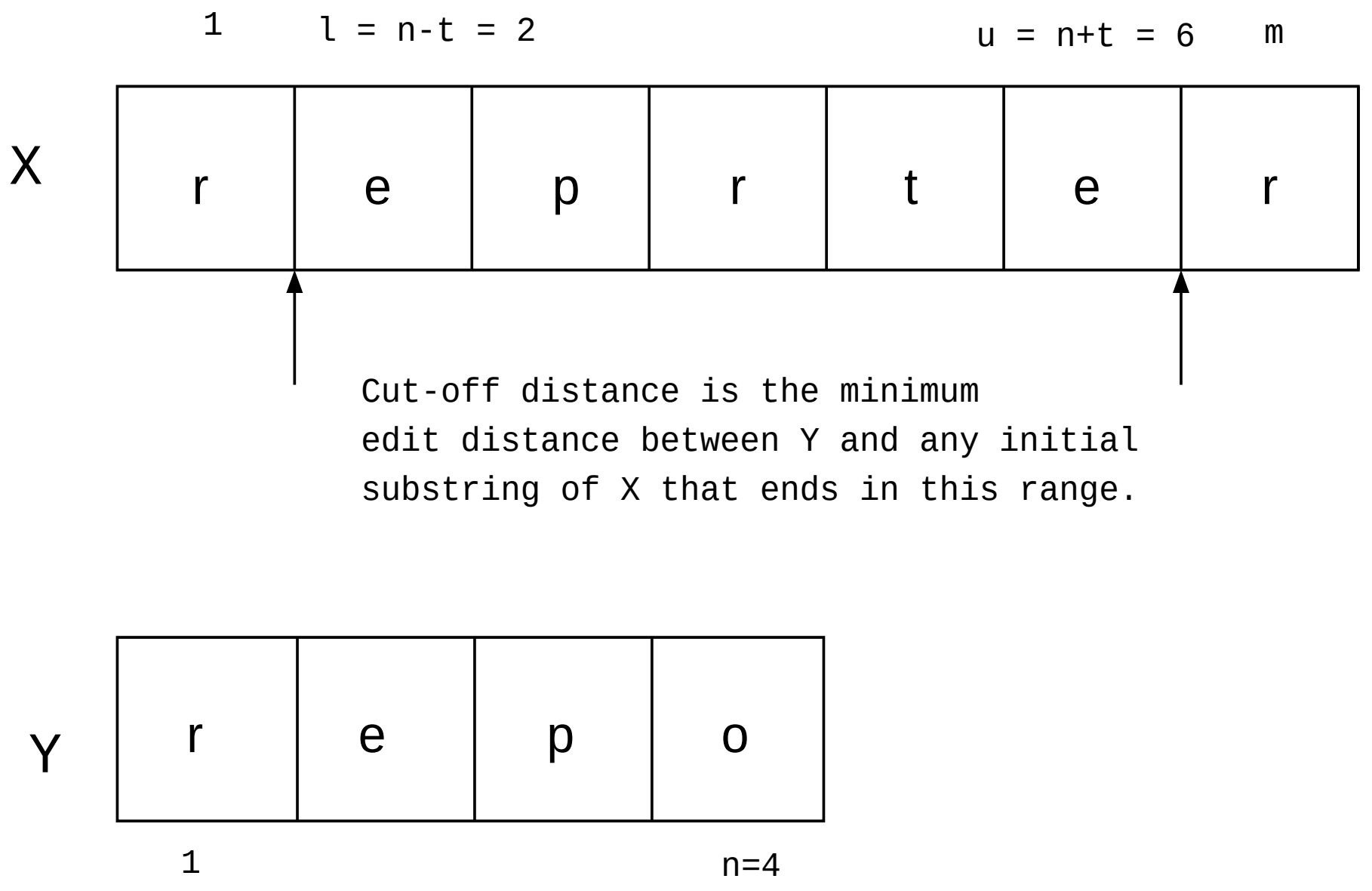

**Figure 2**
The cut-off edit distance.

*edit distance*. The cut-off edit distance measures the minimum edit distance between an initial substring of the incorrect input string, and the (possibly partial) candidate correct string. Let $Y$ be a partial candidate string whose length is $n$, and let $X$ be the incorrect string of length $m$. Let $l = \min(1, n-t)$ and $u = \max(m, n+t)$. The cut-off edit distance $cuted(X[m], Y[n])$ is defined as

$$cuted(X[m], Y[n]) = \min_{l \leq i \leq u} ed(X[i], Y[n]).$$

For example, with $t = 2$:

$$\begin{aligned} cuted(\text{reprter}, \text{repo}) = \min\{ & ed(\text{re}, \text{repo}) = 2, \\ & ed(\text{rep}, \text{repo}) = 1, \\ & ed(\text{repr}, \text{repo}) = 1, \\ & ed(\text{reprt}, \text{repo}) = 2, \\ & ed(\text{reprte}, \text{repo}) = 3\} = 1. \end{aligned}$$

Note that except at the boundaries, the initial substrings of the incorrect string $X$ considered are of length $n-t$ to length $n+t$. Any initial substring of $X$ shorter than $n-t$ needs more than $t$ insertions, and any initial substring of $X$ longer than $n+t$ requires more than $t$ deletions, to at least equal $Y$ in length, violating the edit distance constraint (see Figure 2).

Given an incorrect string $X$, a partial candidate string $Y$ is generated by successively concatenating relevant labels along the arcs as transitions are made, starting with the start state. Whenever we extend $Y$, we check if the cut-off edit distance of $X$ and the partial $Y$, is within the bound specified by the threshold $t$. If the cut-off edit distance

```
/*push empty candidate, and start node to start search */
push((ε, q0))
while stack not empty
    begin
       pop((Y', qi)) /* pop partial surface string Y'
                          and the node */
       for all qj and a such that δ(qi, a) = qj
          begin /* extend the candidate string */
             Y = concat(Y', a) /* n is the current length of Y */
          /* check if Y has deviated too much, if not push */
             if cuted(X[m], Y[n]) ≤ t then push((Y, qj))
          /* also see if we are at a final state */
             if ed(X[m], Y[n]) ≤ t and qj ∈ F then output Y
          end
    end
```

**Figure 3**
Algorithm for error-tolerant recognition

$$\begin{pmatrix} \dots & \dots & \dots & \dots & \dots \\ \vdots & \vdots & \vdots & \vdots & \vdots \\ \dots & H(i-1,j-1) & \dots & \dots & \dots \\ \dots & \dots & H(i,j) & H(i,j+1) & \dots \\ \dots & \dots & H(i+1,j) & H(i+1,j+1) & \dots \\ \vdots & \vdots & \vdots & \vdots & \vdots \\ \dots & \dots & \dots & \dots & \dots \end{pmatrix}$$

**Figure 4**
Computation of the elements of the $H$ matrix.

goes beyond the threshold, the last transition is backed off to the source node (in parallel with the shortening of $Y$) and some other transition is tried. Backtracking is recursively applied when the search can not be continued from that state. If, during the construction of $Y$, a final state is reached without violating the cutoff edit distance constraint, and $ed(X[m], Y[n]) \leq t$ at that point, then $Y$ is a valid correct form of the incorrect input string.[2]

Denoting the states by subscripted $q$'s ($q_0$ being the initial state) and the symbols in the alphabet (and labels on the directed edges) by $a$, we present the algorithm for generating all $Y$'s by a (slightly modified) depth-first probing of the graph in Figure 3. The crucial point in this algorithm is that the cut-off edit distance computation can be performed very efficiently by maintaining a matrix $H$ which is an $m$ by $n$ matrix with element $H(i,j) = ed(X[i], Y[j])$ (Du and Chang, 1992). We can note that the computation of the element $H(i+1, j+1)$ recursively depends on only $H(i,j)$, $H(i,j+1)$, $H(i+1,j)$ and $H(i-1,j-1)$, from the earlier definition of the edit distance (see Figure 4.)

During the depth first search of the state graph of the recognizer, entries in column $n$ of the matrix $H$ have to be (re)computed, only when the candidate string is of length $n$. During backtracking, the entries for the last column are discarded, but the entries in prior columns are still valid. Thus all entries required by $H(i+1, j+1)$, except $H(i, j+1)$, are already available in the matrix in columns $i-1$ and $i$. The computation of $cuted(X[m], Y[n])$ involves a loop in which the minimum is computed. This loop

2 Note that we have to do this check since we may come to other irrelevant final states during the search.

(indexing along column $j+1$) computes $H(i, j+1)$ before it is needed for the computation of $H(i+1, j+1)$.

We now present in Figure 5 a simple example for this search algorithm for a simple finite state recognizer for the regular expression $(aba + bab)^*$, and the search graph for the input string $ababa$. The thick circles from left to right indicate the nodes at which

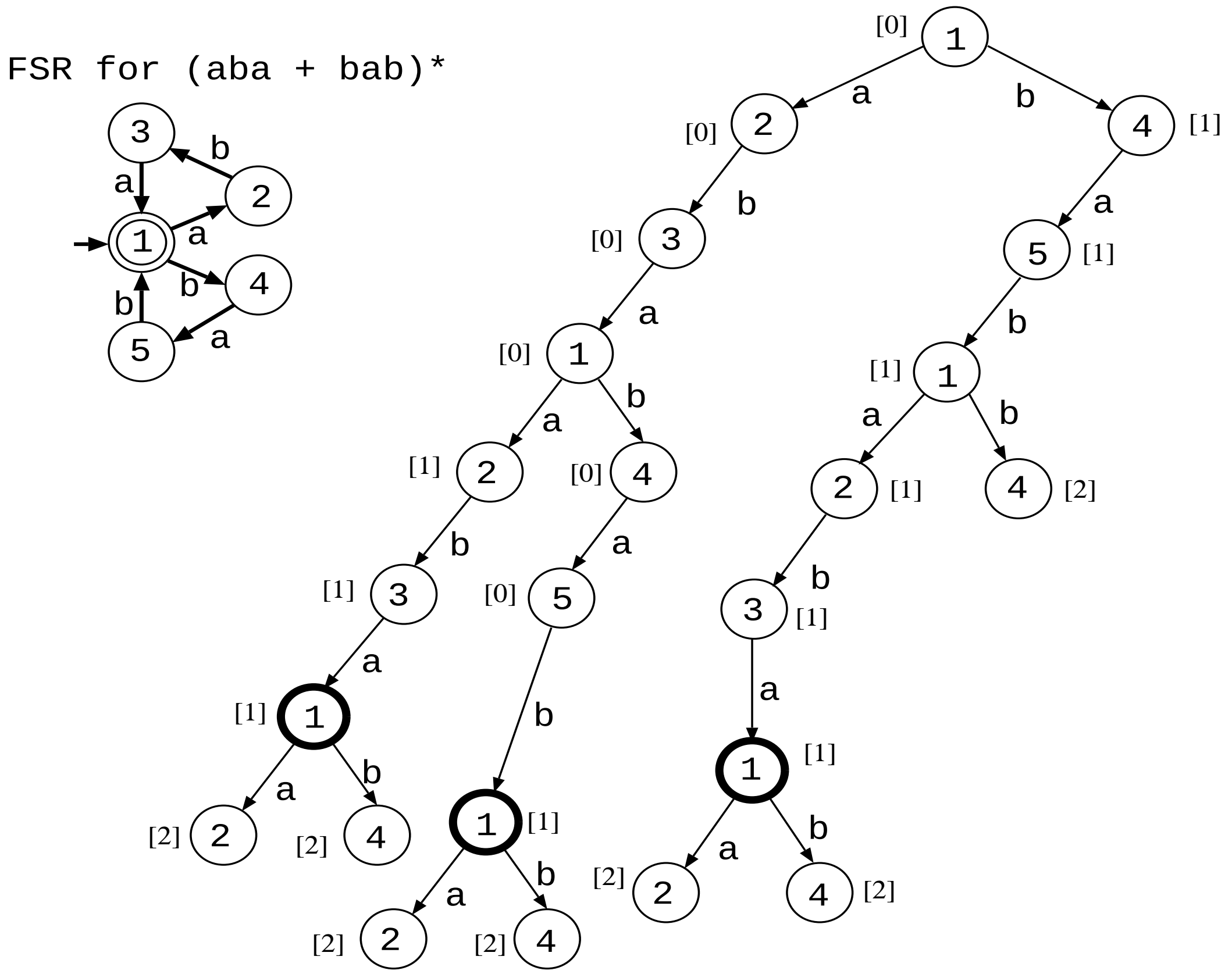

Search graph for matching ababa with threshold 1

Numbers in [ ]'s show the the cut-off edit distance when search reaches that node. States with bold circle indicate successful matches. The strings matching (left-to-right) are abaaba, ababab and bababa.

**Figure 5**
Recognizer for $(aba + bab)^*$ and search graph for $ababa$.

we have the matching strings $abaaba$, $ababab$ and $bababa$, respectively. Prior visits to the final state 1, violate the final edit distance constraint. (Note that the visit order of siblings depend on how one orders the outgoing arcs from a state.)

## 3. Application to Error-tolerant Morphological Analysis

Error-tolerant finite state recognition can be applied to morphological analysis, in which, instead of rejecting a given misspelled form, the analyzer attempts to apply the morphological analysis to forms that are within a certain (configurable) edit distance of the incorrect form. Two-level transducers (Karttunen and Beesley, 1992; Karttunen, Kaplan, and Zaenen, 1992) provide a suitable model for the application of error-tolerant recognition. Such transducers capture all morphotactic and morphographemic phenomena, and alternations in the language in a uniform manner. They can be abstracted as finite state transducers over an alphabet of lexical and surface symbol pairs `l:s`, where either `l` or `s` (but not both) may be the null symbol `0`. It is possible to apply error-tolerant recognition to languages whose word formations employ productive compounding and/or agglutination, and in fact to any language whose morphology is described completely as one (very large) finite state transducer. Full scale descriptions using this approach already exist for a number of languages like English, French, German, Turkish, Korean (Karttunen, 1994).

Application of error-tolerant recognition to morphological analysis proceeds as described earlier. After a successful match with a surface symbol, the corresponding lexical symbol is appended to the output gloss string. During backtracking the candidate surface string and the gloss string are again shortened in tandem. The basic algorithm for this case is given in Figure 6.[3] The actual algorithm is a slightly optimized version of this where transitions with null surface symbols are treated as special during forward and backtracking traversals to avoid unnecessary computations of the cut-off edit distance.

```
/*push empty candidate string, and start node
to start search on to the stack */
push((ε, ε, q0))
while stack not empty
    begin
       pop((surface′, lexical′, qi)) /* pop partial strings
            and the node from the stack */
       for all qj and l : s such that δ(qi, l : s) = qj
          begin /* extend the candidate string */
             surface = concat(surface′, s)
             if cuted(X[m], surface[n]) ≤ t then
                begin
                    lexical = concat(lexical′, l)
                    push((surface, lexical, qj))
                    if ed(X[m], surface[n]) ≤ t and qj ∈ F then
                      output lexical
                end
          end
    end
```

**Figure 6**
Algorithm for error-tolerant morphological analysis.

We can demonstrate error-tolerant morphological analysis with a two–level transducer for the analysis of Turkish morphology. Agglutinative languages such as Turkish, Hungarian or Finnish, differ from languages like English in the way lexical forms are generated. Words are formed by productive affixations of derivational and inflectional affixes to roots or stems like "beads-on-a-string" (Sproat, 1992). Furthermore, roots and affixes may undergo changes due to various phonetic interactions. A typical nominal or a verbal root gives rise to thousands of valid forms which never appear in the dictionary. For instance, we can give the following (rather exaggerated) adverb example from Turkish:

*uygarlaştıramayabileceklerimizdenmişsinizcesine*

3 Note that transitions are now labeled with $l : s$ pairs.

whose root is the adjective *uygar* (civilized).[4] The morpheme breakdown (with morphological glosses underneath) is:[5]

| | | | | | |
|---|---|---|---|---|---|
| *uygar* | *+laş* | *+tır* | *+ama* | *+yabil* | *+ecek* |
| civilized | +AtoV | +CAUS | +NEG | +POT | +VtoA(AtoN) |
| *+ler* | *+imiz* | *+den* | *+miş* | *+siniz* | *+cesine* |
| +3PL | +POSS-1PL | +ABL(+NtoV) | +PAST | +2PL | +VtoAdv |

The portion of the word following the root consists of 11 morphemes each of which either adds further syntactic or semantic information to, or changes the part-of-speech, of the part preceding it. Though most words one uses in Turkish are considerably shorter than this, this example serves to point out some of the fundamental difference of the nature of the word structures in Turkish and other agglutinative languages, from those of languages like English.

Our morphological analyzer for Turkish is based on a lexicon of about 28,000 root words and is a re-implementation of PC-KIMMO (Antworth, 1990) version of the same description (Oflazer, 1993), using Xerox two-level transducer technology (Karttunen and Beesley, 1992). This description of Turkish morphology has 31 two-level rules that implement the morphographemic phenomena such as vowel harmony and consonant changes across morpheme boundaries etc., and about 150 additional rules, again based on the two-level formalism, that fine tune the morphotactics by enforcing sequential and long-distance feature sequencing and co-occurrence constraints, in addition to constraints imposed by standard alternation linkage among various lexicons to implement the paradigms. Turkish morphotactics is circular due to the relativization suffix in the nominal paradigm, and multiple causative suffixes in the verb paradigm. There is also considerable linkage between nominal and verbal morphotactics due to productive derivational suffixes. The minimized finite state transducer constructed by composing the transducers for root lexicons, morphographemic rules and morphotactic constraints, has 32,897 states and 106,047 transitions, with an average fan out of about 3.22 transitions per state (including transitions with null surface symbols). It analyzes a given Turkish lexical form into a sequence of *feature-value tuples* (instead of the more conventional sequence of morpheme glosses) that are used in a number of natural language applications. The Xerox software allows the resulting finite state transducer to be exported in a tabular form which can be imported to other applications.

This transducer has been used as input to an analyzer implementing the error-tolerant recognition algorithm in Figure 6. The analyzer first attempts to parse the input with $t = 0$, and if it fails, relaxes $t$ up to 2, if it can not find any parse with a smaller $t$, and can process about 150 (correct) forms a second on a Sparcstation 10/41.[6,7] Below, we provide a transcript of a run:[8]

```
ENTER WORD > eva
Threshold 0 ... 1 ...
```

4 This is a manner adverb meaning roughly "(behaving) as if you were one of those whom we might not be able to civilize."
5 Glosses in parentheses indicate derivations not explicitly indicated by a morpheme.
6 No attempt was made to compress the finite state recognizer.
7 The Xerox *infl* program working on the proprietary compressed representation of the same transducer can process about 1000 forms/sec on the same platform.
8 The outputs have been slightly edited for formatting. The feature names denote the usual morphosyntactic features. `CONV` denotes derivations to the category indicated by the second token with a suffix or derivation type denoted by the third token, if any.

```
ela     =>  ((CAT ADJ)(ROOT ela))
evla    =>  ((CAT ADJ)(ROOT evla))
ava     =>  ((CAT NOUN)(ROOT av)(AGR 3SG)(POSS NONE)(CASE DAT))
deva    =>  ((CAT NOUN)(ROOT deva)(AGR 3SG)(POSS NONE)(CASE NOM))
eda     =>  ((CAT NOUN)(ROOT eda)(AGR 3SG)(POSS NONE)(CASE NOM))
ela     =>  ((CAT NOUN)(ROOT ela)(AGR 3SG)(POSS NONE)(CASE NOM))
enva    =>  ((CAT NOUN)(ROOT enva)(AGR 3SG)(POSS NONE)(CASE NOM))
reva    =>  ((CAT NOUN)(ROOT reva)(AGR 3SG)(POSS NONE)(CASE NOM))
evi     =>  ((CAT NOUN)(ROOT ev)(AGR 3SG)(POSS NONE)(CASE ACC))
eve     =>  ((CAT NOUN)(ROOT ev)(AGR 3SG)(POSS NONE)(CASE DAT))
ev      =>  ((CAT NOUN)(ROOT ev)(AGR 3SG)(POSS NONE)(CASE NOM))
evi     =>  ((CAT NOUN)(ROOT ev)(AGR 3SG)(POSS 3SG)(CASE NOM))
eza     =>  ((CAT NOUN)(ROOT eza)(AGR 3SG)(POSS NONE)(CASE NOM))
leva    =>  ((CAT NOUN)(ROOT leva)(AGR 3SG)(POSS NONE)(CASE NOM))
neva    =>  ((CAT NOUN)(ROOT neva)(AGR 3SG)(POSS NONE)(CASE NOM))
ova     =>  ((CAT NOUN)(ROOT ova)(AGR 3SG)(POSS NONE)(CASE NOM))
ova     =>  ((CAT VERB)(ROOT ov)(SENSE POS)(MOOD OPT)(AGR 3SG))

ENTER WORD > akıllınnikiler
Threshold 0 ... 1 ... 2 ...



akıllınınkiler =>
            ((CAT NOUN)(ROOT akıl)(CONV ADJ LI)
                  (CONV NOUN)(AGR 3SG) (POSS NONE)(CASE GEN)
                  (CONV PRONOUN REL)(AGR 3PL)(POSS NONE)(CASE NOM))
akıllınınkiler =>
            ((CAT NOUN)(ROOT akıl)(CONV ADJ LI)
                  (CONV NOUN)(AGR 3SG)(POSS 2SG)(CASE GEN)
                  (CONV PRONOUN REL)(AGR 3PL)(POSS NONE)(CASE NOM))
akıllındakiler =>
            ((CAT NOUN)(ROOT akıl)(CONV ADJ LI)
                  (CONV NOUN)(AGR 3SG)(POSS 2SG)(CASE LOC)
                  (CONV ADJ REL)
                  (CONV NOUN)(AGR 3PL)(POSS NONE)(CASE NOM))


ENTER WORD > eviminkinn
Threshold 0 ... 1 ...



eviminkini =>
            ((CAT NOUN)(ROOT ev)(AGR 3SG)(POSS 1SG)(CASE GEN)
                  (CONV PRONOUN REL)(AGR 3SG)(POSS NONE)(CASE ACC))
eviminkine =>
            ((CAT NOUN)(ROOT ev)(AGR 3SG)(POSS 1SG)(CASE GEN)
                  (CONV PRONOUN REL)(AGR 3SG)(POSS NONE)(CASE DAT))
eviminkinin =>
            ((CAT NOUN)(ROOT ev)(AGR 3SG)(POSS 1SG)(CASE GEN)
                  (CONV PRONOUN REL)(AGR 3SG)(POSS NONE)(CASE GEN))

ENTER WORD > teeplerdeki
Threshold 0 ... 1 ...



tepelerdeki =>
            ((CAT NOUN)(ROOT tepe)(AGR 3PL)(POSS NONE)(CASE LOC)
                  (CONV ADJ REL))
teyplerdeki =>
            ((CAT NOUN)(ROOT teyb)(AGR 3PL)(POSS NONE)(CASE LOC)
                  (CONV ADJ REL))
```

```
ENTER WORD > uygarlaştıramadıklarmıızdanmışsınızcasına
Threshold 0 ... 1 ...



uygarlaştıramadıklarımızdanmışsınızcasına =>
            ((CAT ADJ)(ROOT uygar)(CONV VERB LAS)(VOICE CAUS)(SENSE NEG)
                  (CONV ADJ DIK)(AGR 3PL)(POSS 1PL)(CASE ABL)
                  (CONV VERB)(TENSE NARR-PAST)(AGR 2PL)
                  (CONV ADVERB CASINA)(TYPE MANNER))


ENTER WORD > okatulna
Threshold 0 ... 1 ... 2 ...



okutulma =>
            ((CAT VERB)(ROOT oku)(VOICE CAUS)(VOICE PASS)(SENSE NEG)
                  (MOOD IMP)(AGR=2SG))
okutulma =>
            ((CAT VERB)(ROOT oku)(VOICE CAUS)(VOICE PASS)(SENSE POS)
                  (CONV NOUN MA)(TYPE INFINITIVE)
                  (AGR 3SG)(POSS NONE)(CASE NOM))
okutulan =>
            ((CAT VERB)(ROOT oku)(VOICE CAUS)(VOICE PASS)(SENSE POS)
                  (CONV ADJ YAN))
okutulana =>
            ((CAT VERB)(ROOT oku)(VOICE CAUS)(VOICE PASS)(SENSE POS)
                  (CONV ADJ YAN)(CONV NOUN)(AGR 3SG)(POSS NONE)(CASE DAT))
okutulsa => ((CAT VERB)(ROOT oku)(VOICE CAUS)(VOICE PASS)(SENSE POS)
                  (MOOD COND)(AGR 3SG))
okutula =>
            (CAT VERB)(ROOT oku)(VOICE CAUS)(VOICE PASS)(SENSE POS)
                  (MOOD OPT)(AGR 3SG))
```

In an application context, the candidates that are generated by such a morphological analyzer can be disambiguated or filtered to a certain extent by constraint-based tagging techniques e.g., (Oflazer and İlker Kuruöz, 1994; Voutilainen and Tapanainen, 1993) that take into account syntactic context for morphological disambiguation.

## 4. Applications to Spelling Correction

Spelling correction is an important application for error-tolerant recognition. There has been substantial amount of work on spelling correction (see the excellent review by Kukich (Kukich, 1992)). All methods essentially enumerate plausible candidates which resemble the incorrect word, and use additional heuristics to rank the results.[9] However, most techniques assume a word list of all words in the language. These approaches are suitable for languages like English for which it is possible to enumerate such a list. They are not directly suitable or applicable to languages like German, that have very productive compounding, or agglutinative languages like Finnish, Hungarian or Turkish, in which the concept of a word is much larger than what is normally found in a word list. For example, Finnish nouns have about 2000 distinct forms while Finnish verbs have about 12,000 forms (Gazdar and Mellish, 1989, pages 59–60). The case in Turkish is also similar where, for instance nouns may have about 170 different forms, not counting

9 Ranking is dependent on the language, the application, and the error model. It is an important component of the spelling correction problem, but is not addressed in this paper.

the forms for adverbs, verbs, adjectives, or other nominal forms, generated (sometimes circularly) by derivational suffixes. Hankamer (Hankamer, 1989) gives much higher figures (in the millions) for Turkish, presumably by taking into account derivations.

There have been some recent approaches to spelling correction using morphological analysis techniques. Veronis (Veronis, 1988) presents a method for handling quite complex combinations of typographical and phonographic errors, the latter being the kind of errors usually made by language learners using computer-aided instruction. This method takes into account phonetic similarity in addition to standard errors. Aduriz *et al.* (Aduriz *et al.*, 1993) have used a two-level morphology approach to spelling correction in Basque. Their approach uses two-level rules to describe common insertion and deletion errors, in addition to the two-level rules for the morphographemic component. Oflazer and Güzey (Oflazer and Güzey, 1994) have used a two-level morphology approach to spelling correction in agglutinative languages, which has used a coarser morpheme-based morphotactic description instead of the finer lexical/surface symbol approach presented here. The approach presented there essentially generates a valid sequence of the lexical forms of root and suffixes and uses a separate morphographemic component implementing the two-level rules, to derive surface forms. However, the approach presented there is very slow mainly because of the underlying PC-KIMMO morphological analysis and generation system, and can not deal with compounding due its approach for root selection. More recently, Bowden and Kiraz (Bowden and Kiraz, 1995) have used a multi-tape morphological analysis technique for spelling correction in Semitic languages which, in addition to the insertion, deletion, substitution and transposition errors, allows for various language specific errors.

For languages like English where all inflected forms can be included in a word list, the word list can be used to construct a finite state recognizer structured as a standard letter tree recognizer (which has an acyclic graph) as shown in Figure 7, to which error-tolerant recognition can be applied. Furthermore, just as in morphological analysis, transducers for morphological analysis can obviously be used for spelling correction, so one algorithm can be applied to any language whose morphology is described using such transducers. We demonstrate the application of error-tolerant recognition to spelling correction by constructing finite state recognizers in the form of letter tree from large word lists that contain root and inflected forms of words for 10 languages, obtained from a number of resources on the Internet. Table 1 gives statistics about the word lists used. The Dutch, French, German, English (two different lists), and Italian, Norwegian, Swedish, Danish and Spanish word lists contained some or all inflected forms in addition to the basic root forms. The Finnish word list contained unique word forms compiled from a corpus, although the language is agglutinative.

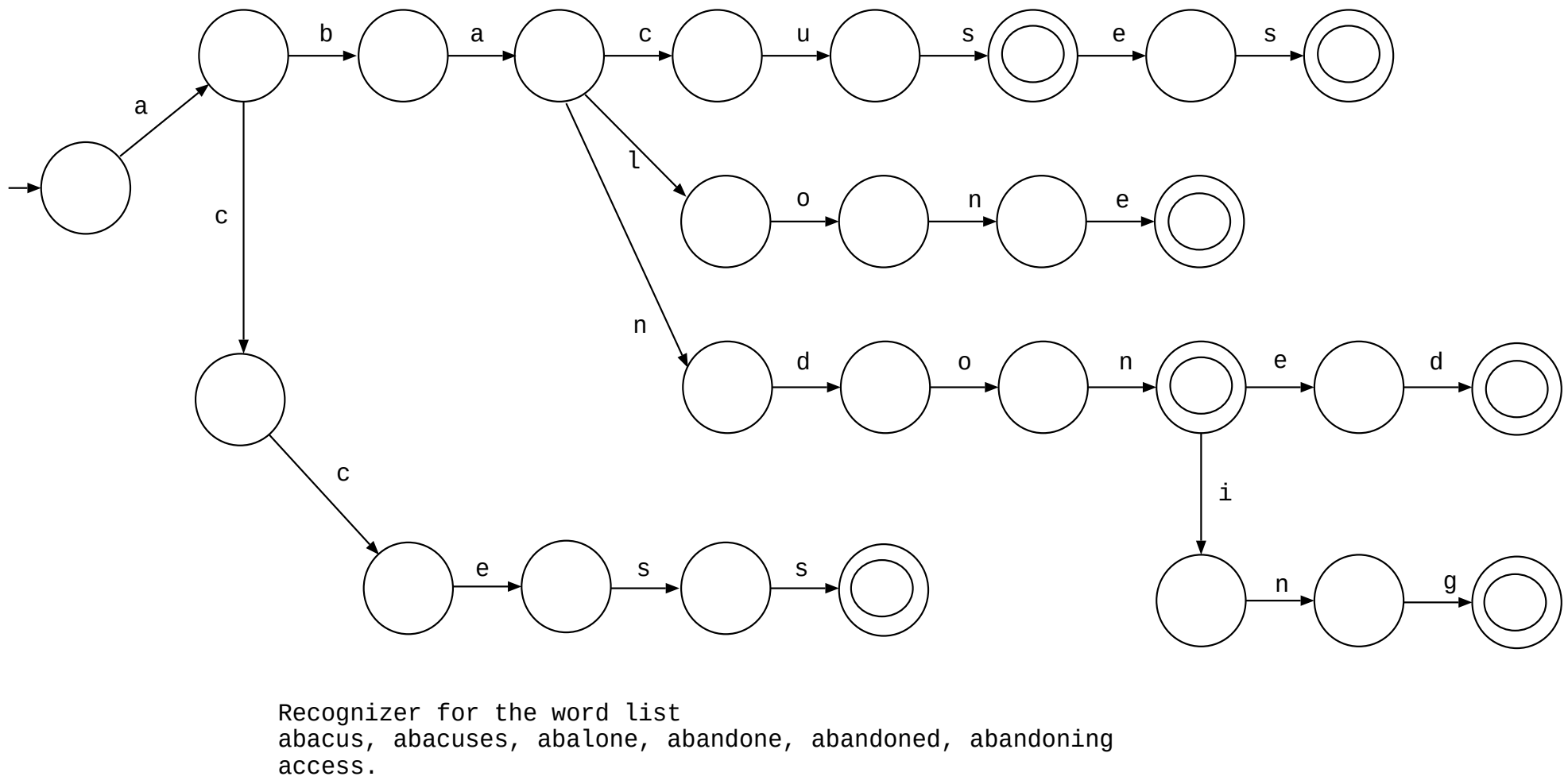

**Figure 7**
A finite state recognizer for a word list.

**Table 1**
Statistics about the language word lists used

| Language | Words | Arcs | Average Word Length | Maximum Word Length | Average Fan-out |
|---|---|---|---|---|---|
| Finnish | 276,448 | 968,171 | 12.01 | 49 | 1.31 |
| English-1 | 213,557 | 741,835 | 10.93 | 25 | 1.33 |
| Dutch | 189,249 | 501,822 | 11.29 | 33 | 1.27 |
| German | 174,573 | 561,533 | 12.95 | 36 | 1.27 |
| French | 138,257 | 286,583 | 9.52 | 26 | 1.50 |
| English-2 | 104,216 | 265,194 | 10.13 | 29 | 1.40 |
| Spanish | 86,061 | 257,704 | 9.88 | 23 | 1.40 |
| Norwegian | 61,843 | 156,548 | 9.52 | 28 | 1.32 |
| Italian | 61,183 | 115,282 | 9.36 | 19 | 1.84 |
| Danish | 25,485 | 81,766 | 10.18 | 29 | 1.27 |
| Swedish | 23,688 | 67,619 | 8.48 | 29 | 1.36 |

For edit distance thresholds, 1, 2, and 3, we selected randomly, 1000 words from each word list and perturbed them by random insertions, deletions, replacements and transpositions so that each misspelled word had the respective edit distance from the correct form. Kukich (Kukich, 1992), citing a number of studies, reports that typically 80% of the misspelled words contain a single error of one of the unit operations, though there are specific applications where the percentage of such errors are lower. Our earlier study of an error model developed for spelling correction in Turkish also indicated similar results (Oflazer and Güzey, 1994).

Tables 2, 3, and 4 present the results from correcting these misspelled word lists for edit distance threshold 1, 2, and 3 respectively. The runs were performed on a Sparcstation 10/41. The second column in these tables gives the average length of the misspelled string in the input list. The third column gives the time in milliseconds to generate *all* solutions, while the fourth column gives the time to find the first solution. The fifth column gives the average number of solutions generated from the given misspelled strings with the given edit distance. Finally, the last column gives the percentage of the search space (that is, the ratio of forward traversed arcs to the total number of arcs) that is searched when generating all the solutions.

### 4.1 Spelling correction for agglutinative word forms

The transducer for Turkish developed for morphological analysis, using the Xerox software, was also used for spelling correction. However, the original transducer had to be simplified into a recognizer for two reasons. For morphological analysis, the concurrent generation of the lexical gloss string requires that occasional transitions with an empty surface symbol be taken, to generate the gloss properly. Secondly, a given surface form can morphologically be interpreted in many ways which is important in morphological processing. In spelling correction, the presentation of only one of such surface forms is sufficient. To remove all empty transitions and analyses with same surface forms from the Turkish transducer, a recognizer recognizing only the surface forms was extracted by using the Xerox tool *ifsm*. The resulting recognizer had 28,825 states and 118,352 transitions labeled with just surface symbols. The average fan-out of the states in this recognizer was about 4. This transducer was then used to perform spelling correction experiments in Turkish.

**Table 2**
Correction Statistics for Threshold 1

| Language | Average Misspelled Word Length | Average Correction Time (msec) | Avg. Time to First Solution (msec) | Average Number of Solutions Found | Average % of Space Searched |
|---|---|---|---|---|---|
| Finnish | 11.08 | 45.45 | 25.02 | 1.72 | 0.21 |
| English-1 | 9.98 | 26.59 | 12.49 | 1.48 | 0.19 |
| Dutch | 10.23 | 20.65 | 9.54 | 1.65 | 0.20 |
| German | 11.95 | 27.09 | 14.71 | 1.48 | 0.20 |
| French | 10.04 | 15.16 | 6.09 | 1.70 | 0.28 |
| English-2 | 9.26 | 17.13 | 7.51 | 1.77 | 0.35 |
| Spanish | 8.98 | 18.26 | 7.91 | 1.63 | 0.37 |
| Norwegian | 8.44 | 16.44 | 6.86 | 2.52 | 0.62 |
| Italian | 8.43 | 9.74 | 4.30 | 1.78 | 0.46 |
| Danish | 8.78 | 14.21 | 1.98 | 2.25 | 1.00 |
| Swedish | 7.57 | 16.78 | 8.87 | 2.83 | 1.57 |
| Turkish (FSR) | 8.63 | 17.90 | 7.41 | 4.92 | 1.23 |

**Table 3**
Correction Statistics for Threshold 2

| Language | Average Misspelled Word Length | Average Correction Time (msec) | Avg. Time to First Solution (msec) | Average Number of Solutions Found | Average % of Space Searched |
|---|---|---|---|---|---|
| Finnish | 11.05 | 312.26 | 162.49 | 13.54 | 1.30 |
| English-1 | 9.79 | 232.56 | 108.69 | 7.90 | 1.51 |
| Dutch | 10.24 | 148.62 | 68.19 | 9.35 | 1.25 |
| German | 12.05 | 169.88 | 96.55 | 3.33 | 1.14 |
| French | 9.88 | 95.07 | 37.52 | 6.99 | 1.44 |
| English-2 | 9.12 | 129.29 | 55.64 | 12.56 | 2.28 |
| Spanish | 8.78 | 125.35 | 48.80 | 10.24 | 2.49 |
| Norwegian | 8.36 | 112.06 | 42.13 | 27.27 | 3.47 |
| Italian | 8.41 | 57.87 | 25.09 | 8.09 | 2.36 |
| Danish | 9.15 | 82.39 | 34.80 | 13.25 | 4.23 |
| Swedish | 7.44 | 90.59 | 16.47 | 36.37 | 6.84 |
| Turkish (FSR) | 8.59 | 164.81 | 57.87 | 55.12 | 11.12 |

**Table 4**
Correction Statistics for Threshold 3

| Language | Average Misspelled Word Length | Average Correction Time (msec) | Avg. Time to First Solution (msec) | Average Number of Solutions Found | Average % of Space Searched |
|---|---|---|---|---|---|
| Finnish | 11.08 | 1217.56 | 561.70 | 157.39 | 3.86 |
| English-1 | 9.73 | 1001.43 | 413.60 | 87.09 | 5.30 |
| Dutch | 10.30 | 610.52 | 256.90 | 71.89 | 4.07 |
| German | 11.82 | 582.45 | 305.80 | 21.39 | 3.14 |
| French | 9.99 | 349.41 | 122.38 | 41.58 | 4.00 |
| English-2 | 9.36 | 519.83 | 194.69 | 97.24 | 6.97 |
| Spanish | 8.90 | 507.46 | 176.77 | 88.31 | 7.79 |
| Norwegian | 8.47 | 400.57 | 125.52 | 199.72 | 8.98 |
| Italian | 8.34 | 198.79 | 66.80 | 55.47 | 6.41 |
| Danish | 9.25 | 228.55 | 47.9 | 97.85 | 8.69 |
| Swedish | 7.69 | 295.14 | 36.89 | 267.51 | 14.70 |
| Turkish (FSR) | 8.57 | 907.02 | 63.59 | 442.17 | 60.00 |

In the first set of experiments three word lists of 1000 words each were generated from a Turkish corpus, and words were perturbed as described before, for error thresholds of 1, 2, and 3 respectively. The results for correcting these words are presented in the last rows (labeled Turkish (FSR)) of the tables above. It should be noted that the percentage of search space searched may not be very meaningful in this case since the same transitions may be taken in the forward direction, more than once.

On a separate experiment which would simulate a real correction application, about 3000 misspelled Turkish words again compiled from a corpus, were processed by successively relaxing the error threshold starting with $t = 1$. Of these set of words, 79.6% had an edit distance of 1 from the intended correct form, while 15.0% had edit distance 2, and 5.4% had edit distance 3 or more. The average length of the incorrect strings was 9.63 characters. The average correction time was 77.43 milliseconds (with 24.75 milliseconds for the first solution). The average number of candidates offered per correction was 4.29, with an average of 3.62% of the search space being traversed, indicating that this is a very viable approach for real applications. For comparison, the same recognizer running as a spell checker ($t = 0$) can process correct forms at a rate of about 500 words/sec.

## 5. Conclusions

This paper has presented an algorithm for error-tolerant finite state recognition which enables a finite state recognizer to recognize strings that deviate mildly from some string in the underlying regular set, along with results of its application to error-tolerant morphological analysis, and candidate generation in spelling correction. The approach is very fast and applicable to any language with a given a list of root and inflected forms, or with a finite state transducer recognizing or analyzing its word forms. It differs from previous error-tolerant finite state recognition algorithms in that it uses a given finite state machine, and is more suitable for applications where the number of patterns (or the finite state machine) is very large and the string to be matched is small.

On the other hand, there are cases where the proposed approach may not be very efficient and may be augmented with language specific heuristics: For instance in spell correction, users (at least in Turkey, as indicated by our error model (Oflazer and Güzey, 1994)) usually replace non-ASCII characters with their nearest ASCII equivalents due to inconveniences such as non-standard keyboards, or having to input such a character using a sequence of keystrokes. For example, in the last spelling correction experiment for Turkish, almost all incorrect forms with edit distance 3 or more, had 3 or more non-ASCII Turkish characters, all of which were rendered with the nearest ASCII version e.g., *yaşgünümüzde* (on our birthday) was written as *yasgunumuzde*. These can surely be found with appropriate edit distance thresholds, but at the cost of generating many more rather distant words. Under these circumstances, one may use language-specific heuristics first, before resorting to error-tolerant recognition, along the lines suggested by morpholgical analysis based approaches (Aduriz *et al.*, 1993; Bowden and Kiraz, 1995).

Although the method described here does not handle erroneous cases where omission of space characters causes joining of otherwise correct forms (such as *inspite of*), such cases may be handled by augmenting the final state(s) of the recognizers with a transition for space characters and ignoring all but one of such space characters in the edit distance computation.

**Acknowledgments**

This research was supported in part by a NATO Science for Stability Grant TU-LANGUAGE. I would like to thank Xerox Advanced Document Systems, and Lauri Karttunen of Xerox Parc and of Rank Xerox Research Centre (Grenoble) for providing us with the two-level transducer development software. Kemal Ülkü and Kurtuluş Yorulmaz of Bilkent University implemented some of the algorithms. I would like to thank the anonymous reviewers for suggestions and comments that contributed to the improvement of the paper in many respects.

**References**

Aduriz, I., *et al.* 1993. A morphological analysis based method for spelling correction. In *Proceedings of the Sixth Conference of the European Chapter of the Association for Computational Linguistics*, Utrecht, The Netherlands, April.

Antworth, Evan L. 1990. *PC-KIMMO: A two-level processor for Morphological Analysis*. Summer Institute of Linguistics, Dallas, Texas.

Bowden, Tanya and George A. Kiraz. 1995. A morphographemic model for error correction in nonconcatenative strings. In *Proceedings of the* $33^{rd}$ *Annual Meeting of the Association for Computational Linguistics*, pages 24–30, Boston, MA.

Damerau, F. J. 1964. A technique for computer detection and correction of spelling errors. *Communications of the Association for Computing Machinery*, 7(3):171–176.

Du, M .W. and S. C. Chang. 1992. A model and a fast algorithm for multiple errors spelling correction. *Acta Informatica*, 29:281–302.

Gazdar, Gerald and Chris Mellish. 1989. *Natural Language Processing in PROLOG, An Introduction to Computational Linguistics*. Addison-Wesley Publishing Company.

Hankamer, Jorge. 1989. Morphological parsing and the lexicon. In W. Marslen-Wilson, editor, *Lexical Representation and Process*. MIT Press.

Hopcroft, John. E. and Jeffrey. D. Ullman. 1979. *Introduction to Automata Theory, Languages, and Computation*. Addison-Wesley Publishing Company, Reading, Massachusetts.

Karttunen, Lauri. 1994. Constructing lexical transducers. In *Proceedings of the* $16^{th}$ *International Conference on Computational Linguistics*, volume 1, pages 406–411, Kyoto, Japan. International Committee on Computational Linguistics.

Karttunen, Lauri and Kenneth. R. Beesley. 1992. Two-level rule compiler. Technical Report, XEROX Palo Alto Research Center.

Karttunen, Lauri, Ronald M. Kaplan, and Annie Zaenen. 1992. Two-level morphology with composition. In *Proceedings of the* $15^{th}$ *International Conference on Computational Linguistics*, volume 1, pages 141–148, Nantes, France. International Committee on Computational Linguistics.

Kukich, Karen. 1992. Techniques for automatically correcting words in text. *ACM Computing Surveys*, 24:377–439.

Myers, Eugene W. and Webb Miller. 1989. Approximate matching of regular expressions. *Bulletin of Mathematical Biology*, 51(1):5–37.

Oflazer, Kemal. 1993. Two-level description of Turkish morphology. In *Proceedings of the Sixth Conference of the European Chapter of the Association for Computational Linguistics (A full version appears in* Literary and Linguistic Computing, *Vol.9 No.2, 1994.)*, Utrecht, The Netherlands, April.

Oflazer, Kemal and Cemalettin Güzey. 1994. Spelling correction in agglutinative languages. In *Proceedings of the* $4^{th}$ *Conference on Applied Natural Language Processing*, pages 194–195, Stuttgart, Germany, October.

Oflazer, Kemal and İlker Kuruöz. 1994. Tagging and morphological disambiguation of Turkish text. In *Proceedings of the* $4^{th}$ *Conference on Applied Natural Language Processing*, pages 144–149, Stuttgart, Germany, October.

Roche, Emmanuel and Yves Schabes. 1995. Deterministic part-of-speech tagging with finite-state transducers. *Computational Linguistics*, 21(2):227–253, June.

Schneider, Mordechay, H. Lim, and William Shoaff. 1992. The utilization of fuzzy sets in the recognition of imperfect strings. *Fuzzy Sets and Systems*, 49:331–337.

Sproat, Richard. 1992. *Morphology and Computation*. MIT Press, Cambridge, MA.

Veronis, Jean. 1988. Morphosyntactic correction in natural language interfaces. In *Proceedings of* $13^{th}$ *International Conference on Computational Linguistics*, pages 708–713. International Committee on Computational Linguistics.

Voutilainen, Atro and Pasi Tapanainen. 1993. Ambiguity resolution in a reductionistic parser. In *Proceedings of the Sixth Conference of the European Chapter of the Association for Computational Linguistics*, pages 394–403, Utrecht, The Netherlands, April.

Wu, Sun and Udi Manber. 1991. Fast text searching with errors. Technical Report TR91–11, Department of Computer Science, University of Arizona.